\pdfoutput=1
\RequirePackage{fix-cm}
\documentclass[11pt,letterpaper]{article}
\usepackage{loopedwam-arxiv}

\hypersetup{
  pdftitle={LoopedWAM: Looped World Action Models for Elastic Inference},
  pdfauthor={Anh Dao, Xinxin Liu, Hang Yu, Jeffrey A Chan Santiago, Yuzhang Shang},
  pdfkeywords={world action models, looped transformers, weight sharing, elastic inference, robot learning}}

\paperlogos{\logoslot{logos/ucf}}
\papertitle{\textcolor{accent}{LoopedWAM}: Looped World Action Models\\ for Elastic Inference}
\paperauthors{%
  \textbf{Anh Dao}\corr\authorsep
  \textbf{Xinxin Liu}\authorsep
  \textbf{Hang Yu}\authorsep
  \textbf{Jeffrey A Chan Santiago}\authorsep
  \textbf{Yuzhang Shang}}
\paperaffiliations{University of Central Florida}
\paperlinks{%
  \linkrow{\faGlobe}{Project page}{https://anhdao69.github.io/LoopedWAM}{https://anhdao69.github.io/LoopedWAM}
  \linkrow{\faGithub}{Code}{https://github.com/anhdao69/LoopedWAM}{https://github.com/anhdao69/LoopedWAM}
  \linkrow{\iconslot{logos/huggingface}{\faCubes}}{Models}{https://huggingface.co/anhdao69/LoopedWAM}{https://huggingface.co/anhdao69/LoopedWAM}
  \linkrow{\faEnvelope[regular]}{Correspondence}{mailto:anh.dao@ucf.edu}{anh.dao@ucf.edu}}
\shorttitle{LoopedWAM: Looped World Action Models for Elastic Inference}
\shortauthors{Dao et al.}

\begin{document}

\makepapertitle
\begin{abstractcard}
World action models (WAMs) couple generative video modeling with robotic action prediction but typically require large video and action transformers, limiting deployment efficiency.
We present \method, a compact WAM that recurrently applies a shared six-block core in each expert, executing 30 transformer layers while storing only 12 blocks per expert.
To account for their asymmetric inference costs, the video and action experts can operate at different depths: visual features are computed once per query, whereas action computation repeats at every sampling step.
We further propose elastic self-distillation, which trains a single elastic checkpoint, \methodE, across action-depth and sampling-step budgets.
With only 0.58B policy parameters and no robot pretraining, \method achieves 96.1\% success on LIBERO, 91.9\% on RoboTwin~2.0, and 52.2\% on RoboCasa-GR1, while using $7.4\times$ less peak GPU memory than Fast-WAM under matched inference measurements.
On LIBERO, \methodE achieves 96.0\% success in 27.6\,ms or 95.2\% in 24.7\,ms on an H100, providing multiple accuracy--latency operating points from a single checkpoint; with two sampling steps and four action loops, it stays within 0.5 points of \method on all three benchmarks.
Ablations demonstrate that weight-shared depth preserves competitive performance, while aggressive action-depth reduction requires explicit budget-aware training.
On a real AgileX PiPER arm, \method succeeds in 71.3\% of 80 trials across four tasks.
\end{abstractcard}

\begin{figure}[H]
\centering
\includegraphics[width=\textwidth]{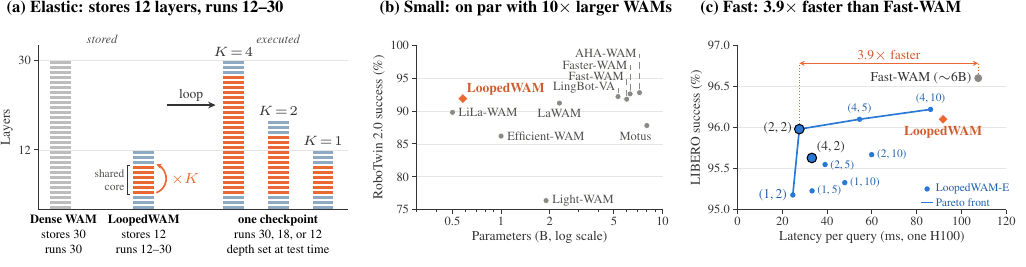}
\caption{\textbf{\method is small, fast, and elastic.}
\textbf{(a)}~Each expert stores 12 blocks and loops its shared 6-block core $K$ times. One checkpoint thus reaches the 30-layer depth of its dense donor with $2.4\times$ fewer parameters, and the depth is chosen at test time.
\textbf{(b)}~RoboTwin~2.0 success against parameters, as reported by each paper. At 0.58B, \method (91.9\%) is the best model below 2B and on par with the $\sim$6B Fast-WAM (91.8\%).
\textbf{(c)}~LIBERO success against latency on one H100 for each setting of action loops $K_a$ and sampling steps $N$ of \methodE. One checkpoint spans 24.7--86.2\,ms at 95.2--96.2\% success, and at $(2,2)$ it is $3.9\times$ faster than Fast-WAM.}
\label{fig:teaser}
\end{figure}

\section{Introduction}
\label{sec:intro}

A practical robot policy must satisfy two competing demands: producing accurate actions from visual observations and language instructions, while operating within the GPU memory and latency constraints of closed-loop control.
Recent world action models (WAMs) have made remarkable progress on the former by leveraging pretrained video diffusion models to learn robotic actions, often through jointly training video and action experts~\cite{bi2026motus, kim2026cosmos, li2026causal, ye2026world, yuan2026fast}.
Drawing on physical priors from web-scale video pretraining, WAMs have achieved competitive performance with vision-language-action models (VLAs) on benchmarks such as LIBERO~\cite{liu2023libero}, RoboTwin~2.0~\cite{chen2025robotwin}, and RoboCasa~\cite{nasiriany2024robocasa}, often without robot pretraining.

However, these gains come at a cost that conflicts with the latter demand.
Most current WAMs carry 2B to 14B parameters and repeatedly execute a deep action expert across multiple denoising steps, whereas closed-loop control calls for a small memory footprint, low latency, and the flexibility to trade accuracy for speed without retraining.
Two natural approaches address different aspects of this challenge, but neither is sufficient on its own.
Sharing weights across depth reduces the number of stored parameters without reducing the sequential computation required at a fixed execution depth.
Conversely, reducing the number of sampling steps lowers latency but may degrade action quality when the policy is not trained for fewer steps.
A practical compact WAM must therefore answer two questions: \textbf{how much capacity must it store, and how much computation must it execute to produce useful actions?}

Answering these questions starts from where a WAM spends its computation.
Fast-WAM~\cite{yuan2026fast} showed that a WAM need not generate future video at test time: the benefit of video modeling comes from joint training, allowing the video expert to encode the current observation once while the action expert generates an action chunk over $N$ sampling steps.
A query therefore requires one video-expert pass and $N$ action-expert passes.
At batch size one, for comparable block architectures, latency is primarily determined by the number of sequentially executed transformer blocks rather than parameter count alone (\cref{sec:efficiency}, \cref{fig:elastic}c).
These observations suggest two complementary opportunities: \textbf{reducing parameter storage through weight sharing and reducing execution cost by allocating computation differently between the two experts.}

\ding{172}~For the first question, we turn to looped, or weight-shared, transformers~\cite{bae2025relaxed, dehghani2018universal, geiping2026scaling, giannou2023looped}, which reach the depth of a deep network with the parameters of a shallow one by repeatedly applying a shared group of blocks.
We bring this idea to world action models with \textbf{\method} (\cref{fig:teaser}a).
Each expert stores 12 blocks organized into a 3-block prelude, a 6-block shared core, and a 3-block coda.
Repeating the shared core four times recovers the 30-layer depth of its Wan2.1 donor~\cite{wan2025wan}.
The resulting policy has 0.58B parameters, $2.4\times$ fewer than an equally deep dense model built from the same donor.
Importantly, this makes execution depth a potential test-time choice without changing the stored parameters.

\ding{173}~For the second question, the asymmetric inference costs of the two experts suggest that their execution depths can be controlled independently.
Given this asymmetry, \method allows the two experts to use different loop counts, with \emph{late alignment} enabling even a single action loop to access the deepest visual features.
However, adjustable execution depth does not guarantee reliable inference at reduced depth: without elastic training, reducing the action expert to a single loop drops LIBERO success to 12.3\% under our ablation training schedule (\cref{fig:elastic}a).
To address this, we introduce \emph{elastic self-distillation}, which trains a single checkpoint, \methodE, to operate across action-loop counts $K_a\in\{1,2,4\}$ and sampling steps $N\in\{2,5,10\}$, while keeping the video expert at full depth.
By matching a frozen full-depth teacher starting from the student's own intermediate states, the student learns to correct errors introduced by reduced computation.

With only 0.58B parameters and no robot pretraining, \method is competitive with WAMs about ten times larger (\cref{fig:teaser}b).
It reaches 91.9\% on RoboTwin~2.0, on par with Fast-WAM~\cite{yuan2026fast} and the best below 2B parameters, 96.1\% on LIBERO, 0.5 points below Fast-WAM under the same protocol, and 52.2\% on RoboCasa-GR1, ahead of StarVLA~\cite{community2026starvla} and DiT4DiT~\cite{ma2026dit4dit}, with $7.4\times$ less peak GPU memory than Fast-WAM.
A single \methodE checkpoint offers nine operating points over a $3.5\times$ latency range (\cref{fig:teaser}c): with two sampling steps and four action loops, it stays within 0.5 points of \method on all three benchmarks, and with two action loops, it reaches 96.0\% on LIBERO in 27.6\,ms per query on one H100, $3.9\times$ faster than Fast-WAM at its default ten steps.
On a real robot arm, \method matches the success of Fast-WAM with $3.4\times$ less memory (\cref{sec:real}).

Our contributions are threefold:
\begin{itemize}
\item \textbf{\method}, a parameter-efficient two-expert WAM that decouples stored parameters from executed depth through weight-shared recurrence, with independently controlled video and action depths and late alignment.
\item \textbf{Elastic self-distillation}, which trains a single checkpoint across action-depth and sampling-step budgets by distilling a frozen full-depth teacher from the student's own intermediate states.
\item \textbf{A systematic evaluation} with dense baselines matched in parameter count or execution depth, controlled ablations, and real-robot experiments, demonstrating the importance of preserving video depth and explicitly training for reduced-depth execution.
\end{itemize}

\section{Related Work}
\label{sec:related}

\subsection{World action models}

Early video-based policies predicted future frames and inferred actions from them~\cite{du2023learning,wu2024unleashing,hu2024video} or diffused video and actions jointly~\cite{Li2025UnifiedVA,zhu2025unified}.
Recent WAMs adapt large pretrained video diffusion transformers~\cite{wan2025wan}: Genie Envisioner~\cite{liao2026genie} and Cosmos Policy~\cite{kim2026cosmos} fine-tune a video model to produce actions, DreamZero~\cite{ye2026world} scales a 14B WAM, Motus~\cite{bi2026motus} and LingBot-VA~\cite{li2026causal} couple video and action experts in a Mixture-of-Transformers (MoT)~\cite{liang2024mixture}, DiT4DiT~\cite{ma2026dit4dit} conditions an action transformer on intermediate video features, and MemoryWAM~\cite{yang2026memorywam} adds persistent memory.
Fast-WAM~\cite{yuan2026fast} showed that the benefit of a WAM comes from video co-training rather than from imagining the future at test time, and several works now cut its cost:
Light-WAM~\cite{li2026light} freezes a 1.3B video backbone, LiLa-WAM~\cite{yang2026lila} and LaWAM~\cite{chen2026lawam} predict compact latent dynamics instead of pixels, Efficient-WAM~\cite{li2026efficient} and Faster-WAM~\cite{zhao2026faster} make future conditioning cheaper, AHA-WAM~\cite{cai2026aha} runs video and action models at different rates, GigaWorld-Policy-0.5~\cite{team2026gigaworld} and SelfWAM~\cite{pan2026selfwam} decode only actions at inference, and Flash-WAM~\cite{akbari2026flash} distills LingBot-VA to one denoising step per modality.
These works shrink or remove components of the model.
\method keeps the full interface, a generative action expert as deep as the video expert, and instead shares depth across loops, which reduces stored parameters and makes the executed depth a test-time choice.

\subsection{Looped and weight-shared transformers}

Universal Transformers~\cite{dehghani2018universal} and ALBERT~\cite{lan2019albert} share weights across depth; looped transformers emulate iterative algorithms~\cite{giannou2023looped,Yang2023LoopedTA,Saunshi2025ReasoningWL} and scale test-time compute~\cite{geiping2026scaling}; relaxed recursive transformers~\cite{bae2025relaxed} turn a pretrained model into a looped one, and Mixture-of-Recursions~\cite{Bae2025MixtureofRecursionsLD} routes tokens to different depths.
Looped World Models~\cite{Lu2026LoopedWM} refine latent states with one shared block, and RD-VLA~\cite{tur2026recurrent} iterates a weight-tied action head until convergence.
Closest to us, ELT~\cite{goyal2026elt} distills intermediate loops from the full loop count for visual generation, and LoopFormer~\cite{jeddi2026loopformer} conditions loops on their step size for elastic-depth language modeling.
All of these loop a single network, and the elastic ones vary only its depth.
\method{} loops both experts of a video--action policy with separate loop counts, since the video expert runs once per query and the action expert once per sampler step.
Late alignment lets any number of action loops read the last video loop, the shared cores start from a non-looped video model, and action depth is elastic jointly with the sampler steps.

\subsection{Fast sampling and elastic computation}

Flow-matching policies~\cite{lipman2022flow,liu2022flow,black2024pi_0} need few integration steps, and step distillation reduces them further~\cite{salimans2022progressivedistillationfastsampling,Song2023ConsistencyM,Frans2024OneSD,Geng2025MeanFF,akbari2026flash}.
Slimmable and once-for-all networks~\cite{Yu2019UniversallySN,Cai2019OnceFA,cai2024flextron} train one set of weights for many widths or depths, and early-exit and adaptive-computation models stop early~\cite{Elbayad2019DepthAdaptiveT,Elhoushi2024LayerSkipEE,Graves2016AdaptiveCT,Banino2021PonderNetLT}.
Elastic self-distillation conditions the student on its step size, as shortcut models do~\cite{Frans2024OneSD}, but distills from the student's own states and over recurrence depth and sampling steps at once, while the video expert, computed once per query, is left unchanged.

\section{Method}
\label{sec:method}

\begin{figure}[t]
\centering
\includegraphics[width=\textwidth]{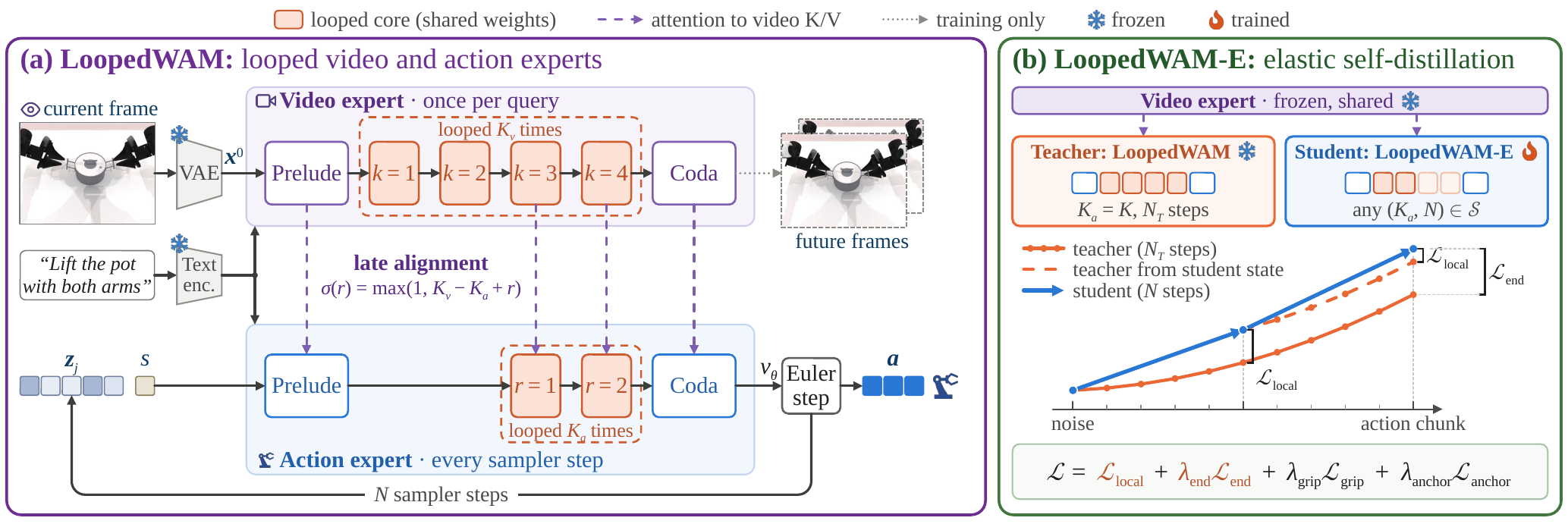}
\caption{\textbf{Method overview.}
\textbf{(a)}~Each expert stores a prelude, a shared core, and a coda, and loops the core $K_v$ times in the video expert and $K_a$ times in the action expert (shown for $K_v{=}4$, $K_a{=}2$). The video expert encodes the current frame once per query and caches its keys and values. At every sampling step, the action expert reads them in a joint attention; late alignment lets action loop $r$ attend to video loop $\sigma(r)$, so even a single action loop sees the deepest visual features. Both experts also cross-attend to frozen text embeddings of the instruction, and future frames are predicted only during training.
\textbf{(b)}~One student learns every setting $(K_a,N)$ in $\mathcal{S}$. The target of each macro step is the frozen teacher's fine Euler map started from the student's \emph{own} state ($\mathcal{L}_{\text{local}}$), and the final chunk is matched to the teacher's chunk from the same noise ($\mathcal{L}_{\text{end}}$).}
\label{fig:method}
\end{figure}

\method separates stored parameter capacity from executed computation depth (\cref{fig:method}).
Both the video and action experts reuse a compact weight-shared transformer core across depth while decoupling their recurrence counts to match their asymmetric inference rates.
Elastic self-distillation then trains a single checkpoint, \methodE, across joint action-depth and sampling-step budgets.

\subsection{Preliminaries: a world action model without test-time imagination}
\label{sec:prelim}

A policy query receives camera images, the proprioceptive state $s$, and a language instruction $\ell$, and returns a chunk of $H$ future actions $\bm{a}\in\mathbb{R}^{H\times D_a}$, the first $H_x$ of which are executed before the next query.
As in Fast-WAM~\cite{yuan2026fast}, a frozen video VAE encodes the images into latent tokens of the current frame $\bm{x}^{0}$ and, during training only, of future frames $\bm{x}^{1:T}$.
A Mixture-of-Transformers~\cite{liang2024mixture} with two experts processes the tokens: a \emph{video expert} for video tokens and a narrower \emph{action expert} for action tokens.
The experts have separate projections, feed-forward layers, and modulation, but their tokens meet in one joint self-attention in every layer, and both cross-attend to frozen text embeddings of $\ell$.

\paragraphx{Attention mask.}
Tokens of the clean current frame attend only to each other, noisy future-frame tokens attend to the current frame and to themselves, and action tokens attend to the current frame and to themselves.
The current frame therefore never depends on actions or future frames and always carries flow time 0, so at test time we drop the future frames, run the video expert \emph{once per query}, and cache its keys and values for the action expert.

\paragraphx{Flow matching.}
Both experts are trained with flow matching~\cite{lipman2022flow,liu2022flow}.
For clean data $\bm{x}$ (actions or future latents), noise $\bm{\epsilon}\sim\mathcal{N}(0,I)$, and time $\tau\in[0,1]$, the model predicts the velocity of $\bm{z}_\tau=(1-\tau)\bm{x}+\tau\bm{\epsilon}$:
\begin{equation}
\mathcal{L}_{\text{flow}} = \mathbb{E}\big[w(\tau)\,\|v_\theta(\bm{z}_\tau,\tau\mid \bm{x}^{0}, s,\ell)-(\bm{\epsilon}-\bm{x})\|^2\big],
\label{eq:fm}
\end{equation}
summed over the action and future-video branches, where $w(\tau)$ is a weighting of the flow time.
At test time the action chunk is integrated from $\tau_0{=}1$ to $\tau_N{=}0$ with $N$ Euler steps, $\bm{z}_{j+1}=\bm{z}_{j}+(\tau_{j+1}-\tau_j)\,v_\theta(\bm{z}_j,\tau_j)$, without classifier-free guidance. A query thus runs the video expert once but the action expert $N$ times, and its latency follows the number of executed layers while its memory follows the number of stored blocks.
We therefore loop a shared core to keep depth with fewer stored blocks (\cref{sec:looped}), let the two experts loop a different number of times (\cref{sec:decouple}), and train one checkpoint to run with fewer action loops and sampling steps (\cref{sec:elastic-method}).

\subsection{Looped experts}
\label{sec:looped}

Each expert stores $L_P{+}L_C{+}L_E$ transformer blocks, split into a prelude $P$ of $L_P$ blocks, a shared core $C$ of $L_C$ blocks, and a coda $E$ of $L_E$ blocks.
With $K$ core loops, an expert maps its input tokens $\bm{u}$ to
\begin{equation}
\bm{h}^{(0)}=P(\bm{u}), \qquad \bm{h}^{(k)}=C\big(\bm{h}^{(k-1)}\big),\ \ k=1,\dots,K, \qquad
\bm{y}=E\big(\bm{h}^{(K)}\big),
\label{eq:loop}
\end{equation}
so it executes $L(K)=L_P+K L_C+L_E$ layers while storing $L_P+L_C+L_E$ blocks: depth grows with $K$ at no cost in parameters.
We choose the partition and the default $K$ so that the executed depth equals that of the donor network the expert is initialized from (\cref{sec:init}).
We keep the loop minimal: there is no input injection, no loop-index embedding, and no outer residual around the core.
At every executed layer the two experts share attention exactly as in a dense MoT.

\subsection{Decoupled video and action recurrence}
\label{sec:decouple}

The two experts run at very different rates: the video expert once per query, the action expert once per sampling step.
\method therefore lets the number of video loops $\Kv$ and action loops $\Ka$ differ (\cref{fig:method}a).
The video expert runs $\Kv$ loops and caches keys and values at each of its $L(\Kv)$ executed layers.
The action expert runs $\Ka$ loops; its prelude and coda attend to the video prelude and coda, and for the core we use \emph{late alignment}: block $b$ of action loop $r\in\{1,\dots,\Ka\}$ attends to block $b$ of video loop
\begin{equation}
\sigma(r)=\max\big(1,\ \Kv-\Ka+r\big),
\label{eq:align}
\end{equation}
so the last action loop always reads the deepest visual features, and $\sigma$ is the identity when $\Ka{=}\Kv$.
A query therefore costs
\begin{equation}
T(\Kv,\Ka,N)\approx T_0 + L(\Kv)\,t_v + N\,L(\Ka)\,t_a ,
\label{eq:cost}
\end{equation}
where $T_0$ is the fixed cost of encoding and I/O, and $t_v$ and $t_a$ are the per-block times of the two experts.
Video depth is paid once per query, action depth $N$ times.
By default \method trains and deploys $\Kv{=}\Ka{=}K$.

\subsection{Initialization from a video diffusion transformer}
\label{sec:init}

Both experts are initialized from a pretrained video diffusion transformer, the \emph{donor}.
The prelude and the coda copy the donor's first $L_P$ and last $L_E$ blocks, and the core copies one contiguous group of $L_C$ early-middle blocks (\cref{sec:ablations} compares other choices).
Where an expert's feed-forward layers are narrower than the donor's, we keep evenly spaced hidden units of the up- and down-projections.
The narrower action expert is derived from the same blocks with the resizing rule of Fast-WAM~\cite{yuan2026fast}.

\subsection{Elastic self-distillation}
\label{sec:elastic-method}

Reducing our fixed-depth model to one action loop causes success to collapse (\cref{sec:exp-elastic}), and fewer sampling steps add integration error.
We therefore train one student, \methodE, to run at every setting in a set $\mathcal{S}$ of action-loop counts $\Ka\le K$ and sampling steps $N$, with the video expert fixed at $\Kv{=}K$ (\cref{fig:method}b).

\paragraphx{Teacher and student.}
The teacher is the trained \method with its $N_T$-step Euler sampler, and it is frozen.
The student starts as a copy of the teacher; its whole action expert is trainable, while the video expert, VAE, text encoder, and proprioceptive projection are frozen and shared with the teacher, so one video pass serves both.
A small zero-initialized step-condition MLP adds an embedding of the step size $d{=}1/N$ to the timestep modulation of the action expert, so the full-depth path initially preserves the teacher's function.

\paragraphx{Targets from the student's own states.}
For a sampled $(\Ka,N)$, student and teacher start from the same noise.
At macro step $j$ the student moves from $\tau_j$ to $\tau_{j+1}=\tau_j-1/N$, and its target is the teacher's Euler map over the same interval \emph{started from the student's current state} $\bm{z}_j$:
\begin{equation}
\mathcal{L}_{\text{local}}=\frac{1}{N}\sum_{j=0}^{N-1}\Big\|\,\hat{v}_j-\mathrm{sg}\Big[\frac{\Phi_{T}(\bm{z}_j;\tau_j{\to}\tau_{j+1})-\bm{z}_j}{\tau_{j+1}-\tau_j}\Big]\Big\|^2 ,
\label{eq:local}
\end{equation}
where $\hat{v}_j$ is the student's velocity at $\bm{z}_j$, $\mathrm{sg}$ stops the gradient, and $\Phi_T$ applies the teacher's steps of size $1/N_T$ that lie in the interval (every macro step ends on the teacher's grid because $N$ divides $N_T$).
Following the teacher from the student's own states, rather than along the teacher's trajectory, exposes the student to its own errors and trains it to correct them (\cref{tab:loss}).

\paragraphx{Endpoint, gripper, and anchor terms.}
An endpoint loss $\mathcal{L}_{\text{end}}$ (mean squared error) matches the $H_x$ executed actions of the student's final chunk to the teacher's chunk from the same noise, $\mathcal{L}_{\text{grip}}$ applies a softplus margin to the gripper command of those actions, and $\mathcal{L}_{\text{anchor}}$ applies \cref{eq:fm} on real actions to the student at $\Ka{=}K$ and $d{=}0$, which keeps the full-depth path close to the data.
The objective is
\begin{equation}
\mathcal{L}=\mathcal{L}_{\text{local}}+\lambda_{\text{end}}\,\mathcal{L}_{\text{end}}+\lambda_{\text{grip}}\,\mathcal{L}_{\text{grip}}+\lambda_{\text{anchor}}\,\mathcal{L}_{\text{anchor}} .
\label{eq:elastic}
\end{equation}
Gradients flow through all loops of a step, but the state is detached at the start of steps $j\le N-m$, so only the last $m$ steps are back-propagated end to end.
Settings are drawn from a balanced sequence that visits each setting of $\mathcal{S}$ once per $|\mathcal{S}|$ updates.
\Cref{alg:elastic} summarizes one update: it draws the next setting from the balanced sequence, integrates the student from noise shared with the teacher, and evaluates the teacher at its fine steps for the local targets and the endpoint.
At deployment, one set of weights serves every setting; an operating point is chosen by setting $\Ka$ and $N$.

\begin{algorithm}[t]
\caption{One update of elastic self-distillation.}
\label{alg:elastic}
\small
\begin{algorithmic}[1]
\Require frozen teacher $\theta_T$ ($\Ka{=}K$, $N_T$ steps); student $\theta_S$; batch $\mathcal{B}$; next $(\Ka,N)\in\mathcal{S}$ from the balanced sequence
\State Encode the current frame once with the frozen video expert ($\Kv{=}K$); cache keys and values
\State Draw shared noise $\bm{z}_0$; $\hat{\bm{a}}_T\gets$ teacher's $N_T$-step chunk from $\bm{z}_0$ \Comment{no grad}
\For{$j=0,\dots,N-1$}
  \If{$j\le N-m$} detach $\bm{z}_j$ \EndIf
  \State $\hat{v}_j\gets v_{\theta_S}(\bm{z}_j,\tau_j;\,d{=}1/N,\Ka)$ \Comment{late alignment, \cref{eq:align}}
  \State $v^{T}_j\gets\big(\Phi_T(\bm{z}_j;\tau_j{\to}\tau_{j+1})-\bm{z}_j\big)/(\tau_{j+1}-\tau_j)$ \Comment{no grad}
  \State $\mathcal{L}_{\text{local}}\mathrel{+}=\|\hat{v}_j-v^{T}_j\|^2/N$;\quad $\bm{z}_{j+1}\gets\bm{z}_j+(\tau_{j+1}-\tau_j)\,\hat{v}_j$
\EndFor
\State $\mathcal{L}_{\text{end}},\mathcal{L}_{\text{grip}}\gets$ first $H_x$ actions of $\bm{z}_N$ vs.\ $\hat{\bm{a}}_T$
\State $\mathcal{L}_{\text{anchor}}\gets$ \cref{eq:fm} on real actions with $\Ka{=}K$, $d{=}0$
\State Update $\theta_S$ with $\mathcal{L}=\mathcal{L}_{\text{local}}+\lambda_{\text{end}}\mathcal{L}_{\text{end}}+\lambda_{\text{grip}}\mathcal{L}_{\text{grip}}+\lambda_{\text{anchor}}\mathcal{L}_{\text{anchor}}$ \Comment{\cref{eq:elastic}}
\end{algorithmic}
\end{algorithm}

\section{Experiments}
\label{sec:experiments}

\subsection{Setup}
\label{sec:setup}

\paragraphx{Benchmarks.}
LIBERO~\cite{liu2023libero} has four suites of ten tasks (Spatial, Object, Goal, Long).
We train one model on all suites and evaluate every task on its 50 initial states with three evaluation seeds, i.e., 6{,}000 episodes per model.
RoboTwin~2.0~\cite{chen2025robotwin} has 50 bimanual tasks; we train one model on 50 clean and 500 randomized demonstrations per task and report success over 100 episodes per task in the clean and the randomized setting.
RoboCasa-GR1~\cite{nasiriany2024robocasa,bjorck2025gr00t} has 24 tabletop tasks for a GR1 humanoid with 29-D actions, evaluated with 50 rollouts per task.

\paragraphx{Model.}
Every model starts from Wan2.1-T2V-1.3B~\cite{wan2025wan} (30 blocks of width 1536), whose VAE and text encoder stay frozen.
Each expert stores $L_P{=}3$, $L_C{=}6$, and $L_E{=}3$ blocks, copied from donor blocks 1--3, 10--15, and 28--30, so with $K{=}4$ it executes 30 layers, the donor's depth.
The video expert keeps width 1536 with 6144 instead of 8960 feed-forward units, the action expert has width 512, and their joint attention has 12 heads of dimension 128.
They hold 479.2M and 105.3M parameters, 584.5M in total (frozen VAE and text encoder excluded).
\method trains and deploys $\Kv{=}\Ka{=}4$ with $N{=}10$ Euler steps, i.e., 300 action-block calls per query, and predicts $H{=}32$ actions per query.
\methodE covers $\mathcal{S}=\{1,2,4\}\times\{2,5,10\}$ for $(\Ka,N)$ with a teacher of $N_T{=}10$ steps; the gripper term $\mathcal{L}_{\text{grip}}$ is used only on LIBERO, whose gripper command is binary.
Two dense controls use the same donor: \textbf{Dense-12} executes the same 12 blocks once (equal parameters), and \textbf{Dense-30} uses all 30 donor blocks (equal depth, $2.4\times$ the parameters).

\paragraphx{Training.}
No model uses robot pretraining, and all are trained on NVIDIA H100 GPUs.
\method is trained on the demonstrations of each benchmark, and \methodE is then distilled from the trained \method of that benchmark.
Ablations use a shorter LIBERO schedule, half the length of the main one, under which \method reaches 94.9\%, and RoboTwin-10, ten RoboTwin tasks trained as on RoboTwin~2.0 (2{,}000 episodes per model).
Detailed training recipes, hyperparameters, and evaluation protocols will be released together with the code.

\paragraphx{Baselines and latency.}
We re-evaluate Fast-WAM~\cite{yuan2026fast}, Faster-WAM~\cite{zhao2026faster}, and Light-WAM~\cite{li2026light} on LIBERO with their released checkpoints; our runs land 0.3--1.4 points below their reported averages (\cref{tab:app-rerun}), so we compare against our runs.
Latency is measured on one H100 at batch size 1, from raw observation to actions, as the mean over 500 queries.

\subsection{Main results}
\label{sec:main}

\begin{table}[t]
\centering
\caption{\textbf{LIBERO} success rate (\%). Results of prior methods are taken from their papers. $^{*}$Re-evaluated by us with the released checkpoints under our protocol (40 tasks $\times$ 50 initial states $\times$ 3 seeds, 6{,}000 episodes), which we also use for our models. Params: policy parameters without the frozen VAE and text encoder. Robot PT: pretraining on robot data; ``--'': not stated. \methodE is one checkpoint run with $K_a$ action loops and $N$ sampling steps.}
\label{tab:libero}
\tabfont
\setlength{\tabcolsep}{8pt}
\begin{tabular}{@{}lccccccc@{}}
\toprule
Method & Params & Robot PT & Spatial & Object & Goal & Long & Avg. \\
\midrule
\tabrows{tables/t_libero_rows}
\bottomrule
\end{tabular}
\end{table}

\begin{table}[t]
\centering
\begin{minipage}[t]{0.575\textwidth}
\centering
\captionof{table}{\textbf{RoboTwin~2.0} (50 tasks, one model) success rate (\%) over 100 episodes per task in each setting. Results of prior methods are taken from their papers.}
\label{tab:robotwin}
\tabfontsmall
\setlength{\tabcolsep}{4.2pt}
\begin{tabular}[t]{@{}lccccc@{}}
\toprule
Method & Params & Robot PT & Clean & Rand. & Avg. \\
\midrule
\tabrows{tables/t_robotwin_rows}
\bottomrule
\end{tabular}
\end{minipage}\hfill
\begin{minipage}[t]{0.395\textwidth}
\centering
\captionof{table}{\textbf{RoboCasa-GR1} (24 tasks) average success rate (\%) over 50 rollouts per task ($^{\dagger}$20; $^{\S}$51; $^{\ddagger}$not stated). \mbox{``--'':} not stated; by task group: \cref{tab:gr1-groups}.}
\label{tab:gr1}
\tabfontsmall
\setlength{\tabcolsep}{4.2pt}
\begin{tabular}[t]{@{}lcc@{}}
\toprule
Method & Params & Avg. \\
\midrule
\tabrows{tables/t_gr1_rows}
\bottomrule
\end{tabular}
\end{minipage}
\end{table}

\paragraphx{LIBERO (\cref{tab:libero}).}
\method reaches 96.1\%, 0.5 points below Fast-WAM$^{*}$ and 2.6 below Faster-WAM$^{*}$, which have $10$--$11\times$ more parameters, and 0.3 points above Light-WAM$^{*}$, all under one protocol.
Both gaps are concentrated in LIBERO-Long (90.1\% vs.\ 93.5\% and 98.1\%), whereas on Object and Goal \method is ahead of Fast-WAM$^{*}$.
\methodE keeps this accuracy with only two sampling steps: 95.6\% at $K_a{=}4$, 96.0\% at $K_a{=}2$.

\paragraphx{RoboTwin~2.0 (\cref{tab:robotwin}).}
\method reaches 91.9\%, with nearly equal success in the clean (92.1\%) and randomized (91.7\%) settings.
This matches Fast-WAM (91.8\%) with a tenth of its parameters and exceeds every model below 2B parameters as well as LaWAM (91.2\%), which is $4\times$ larger and pretrained on robot data; only LingBot-VA, Faster-WAM, and AHA-WAM, all above 5B parameters, are ahead, by 0.3, 0.7, and 0.9 points.

\paragraphx{RoboCasa-GR1 (\cref{tab:gr1}).}
\method reaches 52.2\%, ahead of GR00T-N1.6 (47.6\%), StarVLA-GR00T (47.8\%), Being-H0.7 (49.2\%), and DiT4DiT (50.8\%); the latter three use the same 50-rollout protocol.
LDA-1B (55.4\%), VLAs on 4B backbones (VP-VLA 53.8\%, TwinBrainVLA 54.6\%, ABot-M0 58.3\%), and ACE-Ego-0 (72.8\%, pretrained on 6K hours of robot and human data) score higher, and RoboCasa-GR1 remains our hardest benchmark.

\subsection{Efficiency}
\label{sec:efficiency}

\begin{table}[t]
\centering
\caption{\textbf{Inference cost} in one shared harness (one H100, batch 1, mean over 500 queries).
Mem.: peak GPU memory (VAE included, text encoder not); for -E rows it includes memory retained from previously timed settings (see text). Speed-up: relative to Fast-WAM$^{*}$ with its native 10 steps. Rows without ``-E'' run the undistilled model with $N$ sampling steps; ``-E'' denotes elastic self-distillation (\cref{sec:elastic-method}), applied to Dense-30 over $N\in\{2,5,10\}$ only. LIBERO: protocol of \cref{tab:libero}. RoboTwin~2.0: average of clean and randomized (\cref{tab:robotwin}; prior methods as reported). Light-WAM predicts the action chunk in one forward pass. $^{*}$Re-evaluated by us.}
\label{tab:efficiency}
\tabfont
\setlength{\tabcolsep}{6pt}
\begin{tabular}{@{}lcccccccc@{}}
\toprule
Method & Params & $K_a$ & $N$ & Mem. & Latency & Speed- & LIBERO & RoboTwin \\
       &        &       &     & (GB) & (ms) & up     & (\%) & (\%) \\
\midrule
\tabrows{tables/t_eff_rows}
\bottomrule
\end{tabular}
\end{table}

\Cref{tab:efficiency} compares inference cost in the shared harness.
\method needs $7.4\times$ less GPU memory than Fast-WAM$^{*}$ (2.05 vs.\ 15.25\,GB) and $1.8\times$ less than Dense-30 (3.79\,GB).
At full depth it is only $1.2\times$ faster: both models execute 30 layers per expert in sequence, and at batch size one latency follows the number of executed blocks rather than the number of parameters (\cref{fig:elastic}c); Dense-30, with $2.4\times$ the parameters of \method, is as fast as \method.
The speed comes from executing fewer blocks, which elastic self-distillation allows: two sampling steps instead of ten save 53.0\,ms at four action loops (\cref{tab:elastic9}), and two loops instead of four another 5.6\,ms.
\methodE answers a query in 27.6\,ms at $(K_a,N)=(2,2)$, $3.9\times$ faster than Fast-WAM$^{*}$ at its native ten steps, at 96.0\% LIBERO success, and in 24.7\,ms ($4.4\times$) at $(1,2)$.
Fewer steps give part of this gain: at two steps, Fast-WAM$^{*}$ also reaches 96.0\% (37.1\,ms) and undistilled \method 95.3\% (35.0\,ms).
At matched steps, \methodE at $(2,2)$ equals Fast-WAM$^{*}$ in success but is $1.3\times$ faster with $5.7\times$ less memory, and is $1.2\times$ faster than step-distilled Dense-30-E (95.9\%), as only loops are removable.

\paragraphx{Measurement.}
Latency is measured from raw observation to denormalized actions, including preprocessing, VAE encoding, the video expert, and action sampling.
All methods, including the re-evaluated baselines, are timed in the same inference harness, and all latencies and memory peaks come from one measurement job.
Repeated measurements of the same computation differ by up to 15\% (\method: 91.8\,ms; the 4/4 ablation model: 99.3\,ms; \methodE at $(4,10)$: 86.2\,ms), so we do not interpret smaller differences.
\methodE is timed one operating point after another, so its peak memory (2.58--2.85\,GB) is not directly comparable to that of \method (2.05\,GB): later operating points inherit up to 0.3\,GB from earlier ones, and the two models differ only by the 0.4M-parameter step condition.

\subsection{Elastic inference}
\label{sec:exp-elastic}

\begin{figure}[t]
\centering
\includegraphics[width=\textwidth]{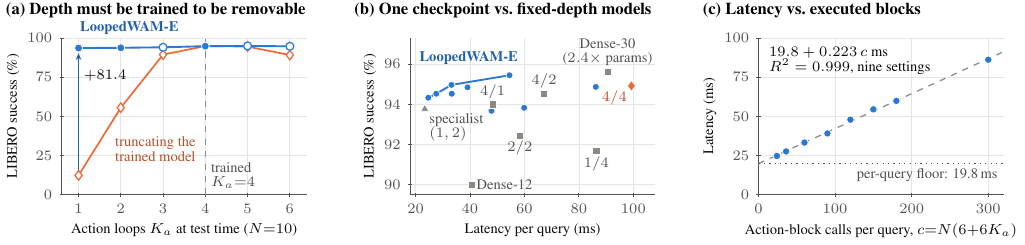}
\caption{\textbf{Elastic inference} (LIBERO, shorter ablation schedule; latency: one H100, batch 1).
\textbf{(a)}~Running fewer action loops than trained collapses the success of the fixed $K_v/K_a{=}4/4$ model (orange), and more loops lower it as well, whereas the elastic student keeps it at every loop count. Filled markers: loop counts used in training; hollow: not used (at $K_a{=}4$ both models score 94.9\%).
\textbf{(b)}~The nine settings of one elastic checkpoint (blue, Pareto frontier as a line) against models trained separately at fixed loop counts $K_v/K_a$ (the $4/4$ model in orange; all fixed models use $N{=}10$), the 12- and 30-layer dense controls, and a specialist distilled only for $(K_a,N)=(1,2)$.
\textbf{(c)}~Latency grows linearly with the number of executed action blocks; the intercept is the per-query cost of encoding, the video expert, and I/O.}
\label{fig:elastic}
\end{figure}

\begin{table}[t]
\centering
\caption{\textbf{All nine operating points of \methodE} on LIBERO (success \%, 6{,}000 episodes each). Calls: action-block executions per query, $N(6{+}6K_a)$. Latency and peak memory: one H100, batch 1. Shaded: the two settings reported in the main tables. Keeping all nine settings loaded at once needs 3.18\,GB. Gain: change from the student distilled from the shorter-schedule model at the same setting (\cref{tab:app-elastic-short}), paired over the same 6{,}000 episodes; significant at the 5\% level in seven of the nine settings (permutation test, unadjusted).}
\label{tab:elastic9}
\tabfont
\setlength{\tabcolsep}{5.5pt}
\begin{tabular}{@{}ccccccccccc@{}}
\toprule
$K_a$ & $N$ & Calls & Latency (ms) & Mem.\ (GB) & Spatial & Object & Goal & Long & Avg. & Gain \\
\midrule
\tabrows{tables/a_elastic9_rows}
\bottomrule
\end{tabular}
\end{table}

\begin{table}[t]
\centering
\caption{\textbf{\methodE on RoboTwin~2.0 and RoboCasa-GR1}: all nine settings of each benchmark's student (success \%; RoboTwin~2.0: 100 episodes per task in each setting; RoboCasa-GR1: 50 rollouts per task). Calls: action-block executions per query. Shaded: the settings of the main tables.}
\label{tab:elastic-rt-gr1}
\tabfont
\setlength{\tabcolsep}{7pt}
\begin{tabular}{@{}ccccccc@{}}
\toprule
 & & & \multicolumn{3}{c}{RoboTwin~2.0} & RoboCasa-GR1 \\
\cmidrule(lr){4-6}\cmidrule(l){7-7}
$K_a$ & $N$ & Calls & Clean & Rand. & Avg. & Avg. \\
\midrule
\tabrows{tables/a_elastic_rt_gr1_rows}
\bottomrule
\end{tabular}
\end{table}

\paragraphx{One checkpoint, nine operating points.}
\Cref{tab:elastic9} lists every setting of \methodE.
Its nine settings span 24.7--86.2\,ms per query ($3.5\times$), while LIBERO success stays between 95.2\% and 96.2\% (\cref{fig:teaser}c); at full depth it matches its teacher (96.2\% vs.\ 96.1\%).
At equal compute, depth beats steps: with 60 action-block calls each, $(K_a,N)=(4,2)$ is 0.4, 2.1, and 5.0 points better than $(1,5)$ on the three benchmarks (\cref{tab:elastic9,tab:elastic-rt-gr1}).
Latency is linear in the number of action-block calls (\cref{fig:elastic}c): an additional sampling step costs 2.90, 4.04, and 6.63\,ms at $K_a=1$, 2, and 4.

\paragraphx{One recipe, three benchmarks (\cref{tab:libero,tab:gr1,tab:elastic-rt-gr1}).}
At two steps and four action loops, a fifth of the teacher's action-block calls, the student stays within 0.5 points of its teacher on all three benchmarks.
Fewer loops cost more on the harder benchmarks: at two steps, one loop loses 2.6 points on RoboTwin~2.0 and 9.0 on RoboCasa-GR1 but only 0.9 on LIBERO.
On RoboCasa-GR1, the four-loop settings reach 51.7--53.5\% and the two-loop settings 50.8--51.6\%, all inside the $\pm$2.8-point 95\% interval of one 1{,}200-episode evaluation around the teacher's 52.2\%, whereas one action loop loses 4.8--9.0 points; the same ordering by loop count holds on RoboTwin~2.0 at every number of sampling steps, where four loops stay within 0.7 points of the teacher and one loop loses 2.4--2.6.
Which settings are worth running thus depends on the task, and one checkpoint covers them all.

\paragraphx{Depth must be trained to be removable (\cref{fig:elastic}a).}
Without elastic training, running the model with fewer action loops than it was trained with collapses: one loop gives 12.3\% and two give 55.6\%.
At the same settings the elastic student keeps 93.7\% and 93.8\%.
It also holds at loop counts it was never trained with: 94.2\%, 95.0\%, and 94.8\% at three, five, and six loops, where the fixed model gets 89.5\%, 94.6\%, and 89.2\% (\cref{tab:app-naive}).

\paragraphx{One checkpoint replaces many models (\cref{fig:elastic}b).}
On the ablation schedule, the elastic checkpoint is both faster and more accurate than every model trained separately with fixed loop counts and than Dense-12; only Dense-30, with $2.4\times$ the parameters, is not dominated.
A specialist distilled only for $(1,2)$ with the same recipe is statistically indistinguishable from the elastic checkpoint at that setting ($-0.6$ points, 95\% CI $-1.3$ to $+0.1$), so training for nine settings has no measurable cost at the fastest one.
As the student is trained further on top of its teacher, while the fixed models are not, this compares deployable models, not training budgets.

\paragraphx{A better teacher gives a better student (\cref{tab:elastic9}).}
Distilling from the main \method (96.1\%) instead of the shorter-schedule model (94.9\%) improves all nine settings, by 0.7--1.8 points (mean 1.1, paired over the same episodes), so elastic self-distillation inherits gains of the teacher.
The gains are largest on LIBERO-Long (2.16 points on average over the nine settings).

\subsection{Ablations}
\label{sec:ablations}

\begin{table}[t]
\centering
\caption{\textbf{Ablations} on the shorter schedule (success, \%; LIBERO: 6{,}000 episodes per model; RoboTwin-10 (RT-10): 2{,}000 episodes per model). $K_v/K_a$: video and action loops; $\Delta$: change from the default (shaded). Per suite: \cref{sec:app-ablations}.}
\label{tab:ablations}
\tabfontsmall
\begin{minipage}[t]{0.40\textwidth}
\centering
\setlength{\tabcolsep}{3.4pt}
\begin{tabular}[t]{@{}lccccc@{}}
\multicolumn{6}{@{}l}{\textit{(a) Depth and loop allocation}} \\
\toprule
Model & $K_v$ & $K_a$ & Params & LIBERO & RT-10 \\
\midrule
\tabrows{tables/t_abl_loops_rows}
\bottomrule
\end{tabular}
\end{minipage}\hfill
\begin{minipage}[t]{0.32\textwidth}
\centering
\setlength{\tabcolsep}{3.8pt}
\begin{tabular}[t]{@{}lcc@{}}
\multicolumn{3}{@{}l}{\textit{(b) Initialization of the core}} \\
\toprule
Wan2.1 blocks & LIBERO & $\Delta$ \\
\midrule
\tabrows{tables/t_abl_init_rows}
\bottomrule
\end{tabular}
\end{minipage}\hfill
\begin{minipage}[t]{0.26\textwidth}
\centering
\setlength{\tabcolsep}{3.8pt}
\begin{tabular}[t]{@{}lccc@{}}
\multicolumn{4}{@{}l}{\textit{(c) Partition $P/C/E$}} \\
\toprule
$P/C/E$ & $K$ & LIBERO & $\Delta$ \\
\midrule
\tabrows{tables/t_abl_part_rows}
\bottomrule
\end{tabular}
\end{minipage}
\end{table}

\paragraphx{Looping recovers most of the benefit of depth (\cref{tab:ablations}a).}
Executing the 12 stored blocks once (Dense-12) gives 90.0\% on LIBERO; looping them to 30 layers gives 94.9\%, and storing 30 distinct blocks (Dense-30) gives 95.6\%.
Looping thus recovers 87\% of the gap between the two dense controls on LIBERO and gains 4 points over \mbox{Dense-12} on RoboTwin-10, without adding parameters.

\paragraphx{Video depth is worth more than action depth.}
At equal parameters, placing the loops in the video expert (4/1) beats placing them in the action expert (1/4) by 2.3 points on LIBERO, at $1.8\times$ lower latency, because video loops are paid once per query (\cref{eq:cost}).
Elastic self-distillation exploits this asymmetry: it keeps four video loops and adjusts only action loops and steps.
On \mbox{RoboTwin-10} the 1.4-point gap is within evaluation noise, but all three models with four video loops lead 1/4 and 2/2 (\cref{sec:app-rt10}).

\paragraphx{Late alignment.}
With one action loop, letting it attend to a learned mixture of all video loops (KV-mix) instead of the deepest one changes LIBERO success by $+$0.45 points, within evaluation noise, at 5\% more latency (\cref{tab:app-loops}); we keep the parameter-free late alignment, which also applies unchanged to every $K_a$ of the elastic student.

\paragraphx{Initialization (\cref{tab:ablations}b).}
A pretrained core matters most: a randomly initialized core with pretrained prelude and coda loses 12.7 points, more than training without any video pretraining ($-$9.2).
The core is best copied from one contiguous group of early-middle donor blocks; later blocks ($-$1.8) or every fourth block ($-$3.4) are worse, and every alternative loses most on LIBERO-Long (\cref{tab:app-init}).

\paragraphx{Partition (\cref{tab:ablations}c).}
With 12 stored blocks and 30 executed layers, the balanced 3/6/3 partition looped four times is best.
A coda is essential (removing it costs 4.6 points), a prelude much less (0.5).
A wider core looped fewer times (1/9/2, $K{=}3$) or a narrower one looped more (4/3/5, $K{=}7$) loses 3.1 and 4.1 points.
Because these partitions also draw on different donor blocks, we read them as joint effects of partition and donor choice.

\begin{table}[t]
\centering
\caption{\textbf{Distillation objective} (success \%). Each row distills \methodE again with one change to \cref{eq:elastic} and otherwise the same teacher, initialization, data order, noise, and sequence of settings. \emph{Teacher trajectory}: every student step, and the local target that supervises it, starts from the teacher's state at $\tau_j$ instead of the student's. LIBERO: students of the main \method, 6{,}000 episodes per setting; Avg.~over all nine settings; $\Delta$: change of that average, paired over the same 54{,}000 episodes. RoboTwin-10: students of the 4/4 model, 2{,}000 episodes per setting; Avg.~over the three settings shown; $\Delta$: unpaired difference; RoboTwin's grippers are continuous, so its objective has no $\mathcal{L}_{\text{grip}}$.}
\label{tab:loss}
\tabfontsmall
\setlength{\tabcolsep}{4.6pt}
\begin{tabular}{@{}lcccccccccc@{}}
\toprule
 & \multicolumn{5}{c}{LIBERO} & \multicolumn{5}{c}{RoboTwin-10} \\
\cmidrule(lr){2-6}\cmidrule(l){7-11}
Objective & $(2,2)$ & $(4,2)$ & $(1,10)$ & Avg. & $\Delta$ & $(2,2)$ & $(4,2)$ & $(1,10)$ & Avg. & $\Delta$ \\
\midrule
\tabrows{tables/a_loss_rows}
\bottomrule
\end{tabular}
\end{table}

\paragraphx{Distillation objective.}
\Cref{tab:loss} changes one part of the objective per run.
\emph{Teacher trajectory} keeps all four terms but evaluates every student step at the teacher's state at that time (teacher forcing), so the student never visits its own states during training; its endpoint for $\mathcal{L}_{\text{end}}$ and $\mathcal{L}_{\text{grip}}$ is the sum of these steps from the shared noise.
LIBERO students are distilled from the main \method and evaluated at all nine settings, and each difference is paired over the same episodes.
RoboTwin-10 students are distilled from the 4/4 RoboTwin-10 model and evaluated at three settings.
Every change lowers the average success, by 0.15--0.33 points on LIBERO and by 0.40--1.20 points on RoboTwin-10, with the teacher-trajectory variant showing the largest drop on RoboTwin-10.

\subsection{Real-world experiments}
\label{sec:real}

\begin{figure}[t]
\centering
\includegraphics[width=\textwidth]{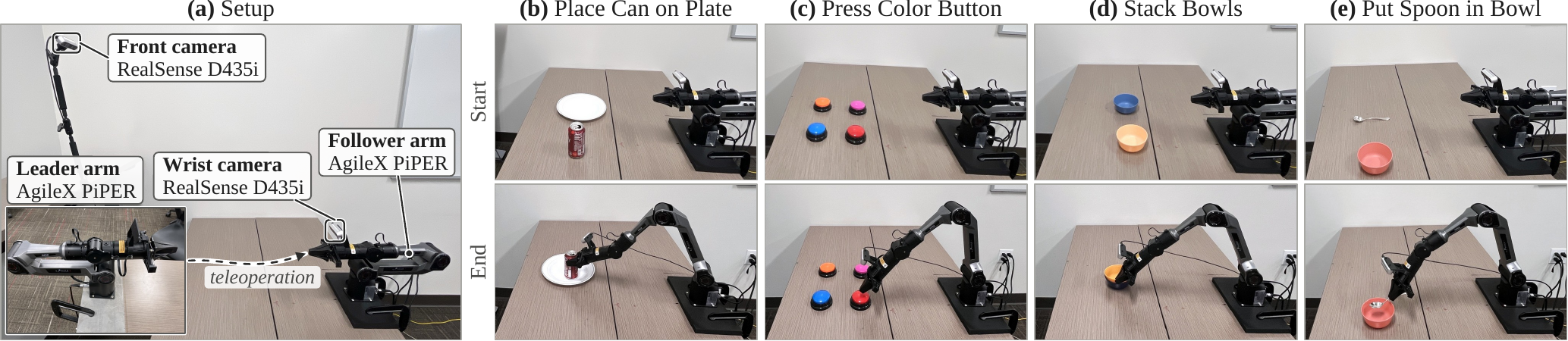}
\caption{\textbf{Real-world setup and tasks.} (a) The follower AgileX PiPER arm with a front and a wrist camera; demonstrations are collected by teleoperation with a second PiPER as the leader arm. (b--e) Start (top) and end (bottom) states of the four tasks. Instructions: ``pick up the can and place it on the plate'', ``press the \{blue, orange, pink, red\} button'', ``stack the bowls'', ``put the spoon in the bowl''.}
\label{fig:real}
\end{figure}

\begin{table}[t]
\centering
\caption{\textbf{Real-robot results} (successful trials out of 20 per task and method, randomized object poses). All models are trained on the same 100 demonstrations per task. $^{*}$Fine-tuned and evaluated by us. Latency is measured on the robot computer (RTX PRO 6000).}
\label{tab:real}
\tabfont
\setlength{\tabcolsep}{8pt}
\begin{tabular}{@{}lcccc@{}}
\toprule
 & & & \multicolumn{2}{c}{\methodE $(K_a,N)$} \\
\cmidrule(l){4-5}
Task & Fast-WAM$^{*}$ & \method & $(4,2)$ & $(2,2)$ \\
\midrule
Place Can on Plate & 14/20 & 14/20 & 14/20 & 13/20 \\
Press Color Button & 18/20 & 18/20 & 17/20 & 17/20 \\
Stack Bowls        & 14/20 & 14/20 & 14/20 & 14/20 \\
Put Spoon in Bowl  & 11/20 & 11/20 & 10/20 & 10/20 \\
\midrule
Total              & 57/80 & 57/80 & 55/80 & 54/80 \\
Success (\%)       & 71.3  & 71.3  & 68.8  & 67.5  \\
Latency (ms)       & 150   & 160   & 58    & 46    \\
Mem (GB)           & 12.7  & 3.7   & 3.3   & 3.3   \\
\bottomrule
\end{tabular}
\end{table}

\begin{figure}[p]
\centering
\newcommand{\rrimg}[1]{\includegraphics[width=0.152\textwidth]{fig/real_rollouts/#1}}
\newcommand{\rrlab}[1]{\rotatebox{90}{\scriptsize\sffamily #1}}
\setlength{\tabcolsep}{0.6pt}\renewcommand{\arraystretch}{0.5}
\begin{tabular}{@{}>{\centering\arraybackslash}m{0.022\textwidth}*{6}{>{\centering\arraybackslash}m{0.152\textwidth}}@{}}
\rrlab{Can} & \rrimg{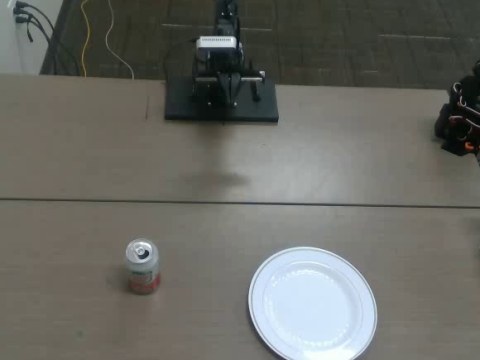} & \rrimg{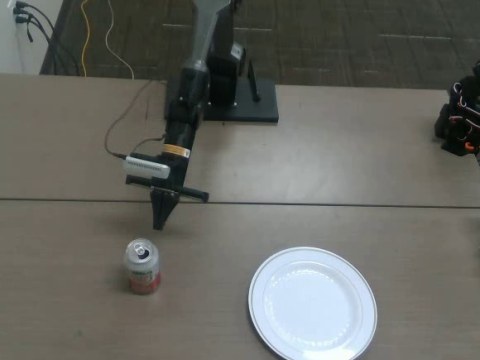} & \rrimg{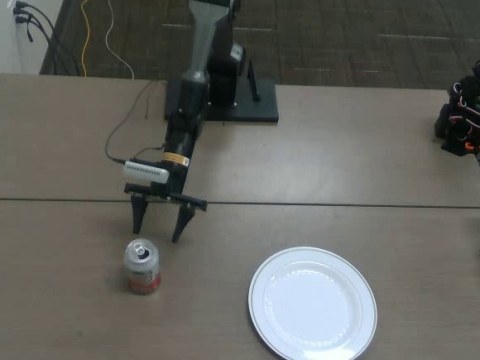} & \rrimg{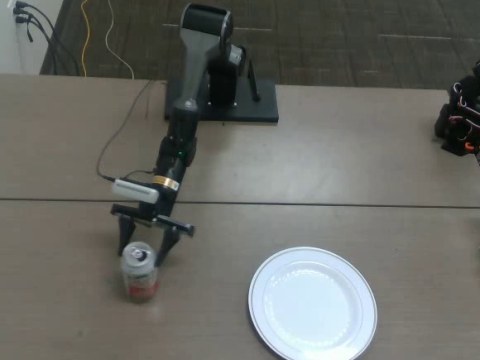} & \rrimg{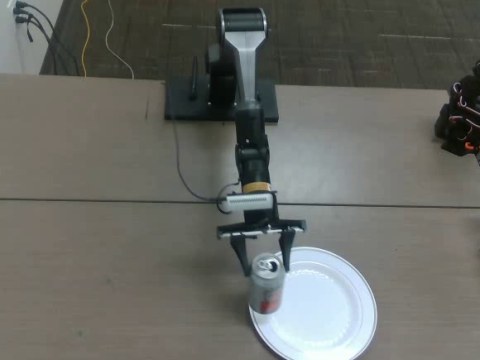} & \rrimg{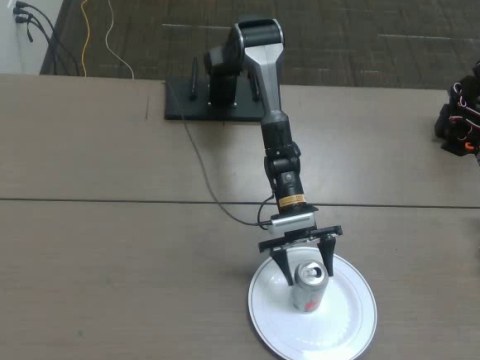} \\
\rrlab{Buttons} & \rrimg{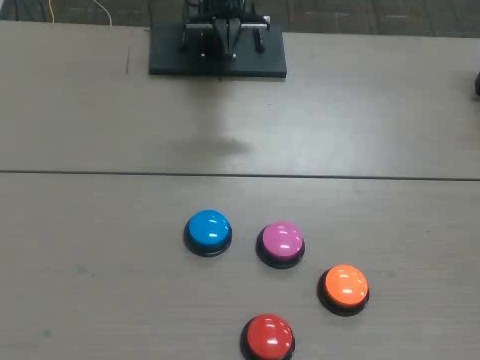} & \rrimg{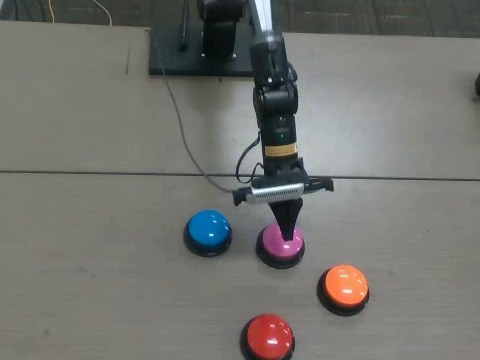} & \rrimg{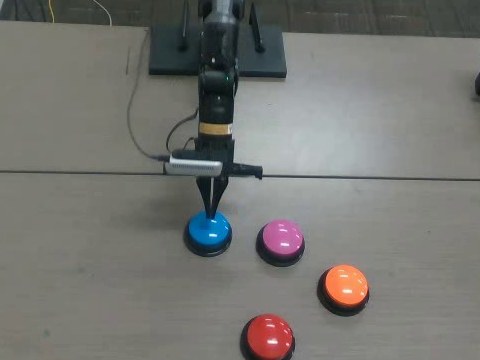} & \rrimg{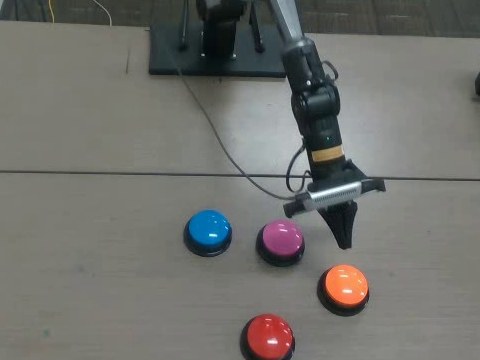} & \rrimg{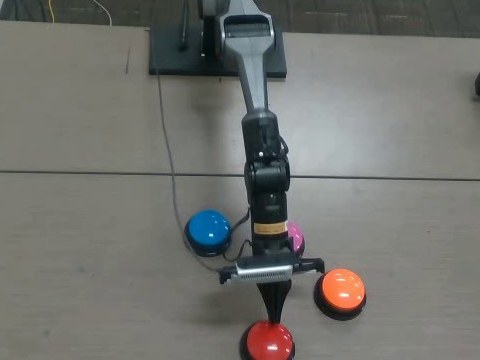} & \\[-1pt]
 & \scriptsize start & \scriptsize ``pink'' & \scriptsize ``blue'' & \scriptsize ``orange'' & \scriptsize ``red'' & \\[3pt]
\rrlab{Bowls} & \rrimg{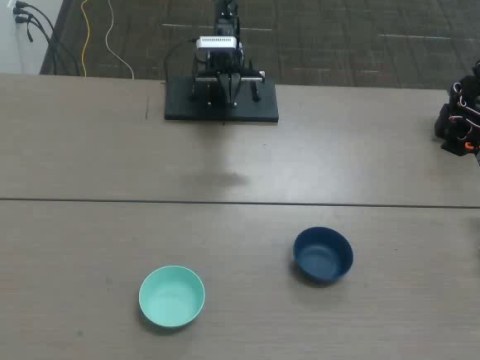} & \rrimg{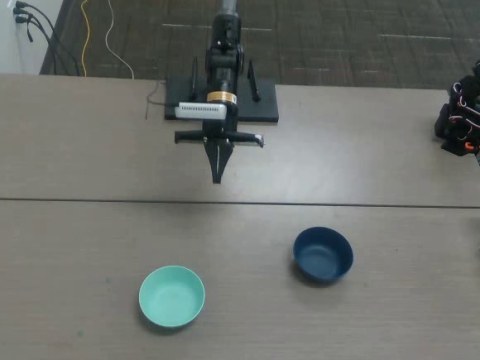} & \rrimg{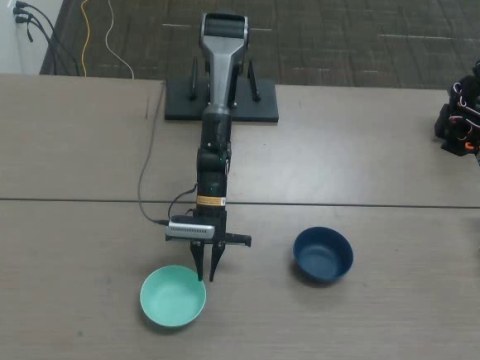} & \rrimg{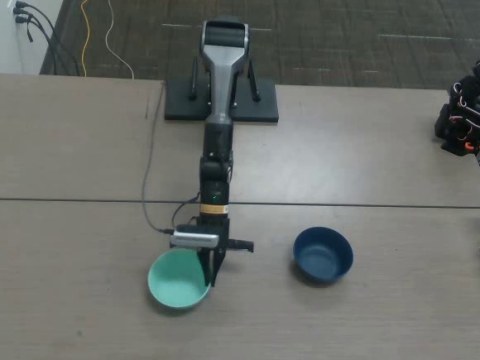} & \rrimg{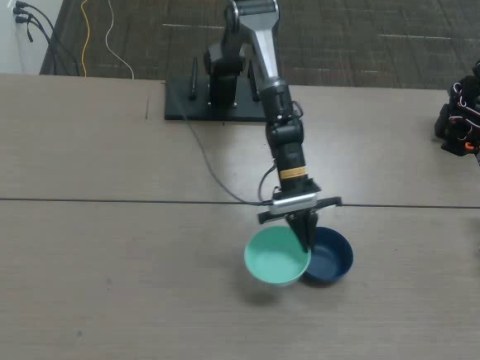} & \rrimg{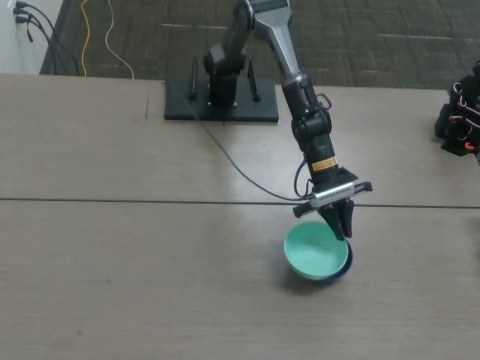} \\
\rrlab{Spoon} & \rrimg{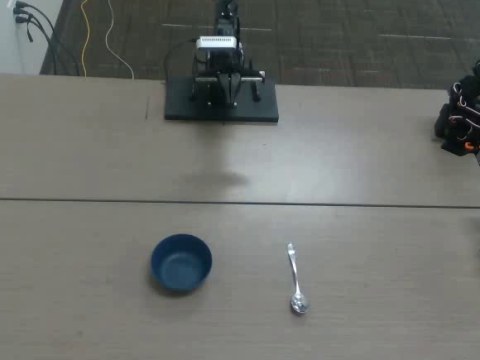} & \rrimg{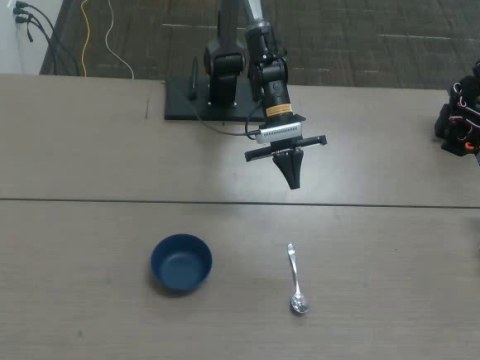} & \rrimg{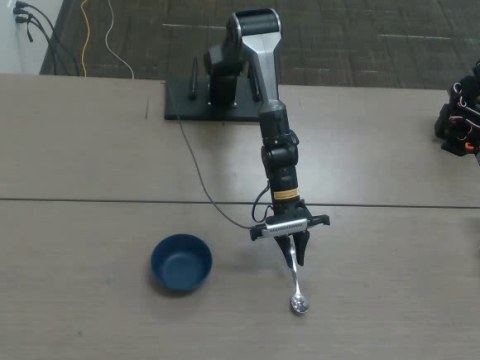} & \rrimg{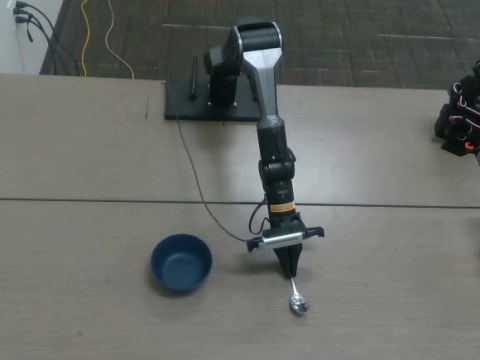} & \rrimg{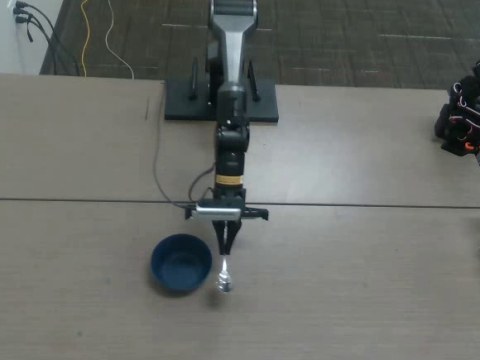} & \rrimg{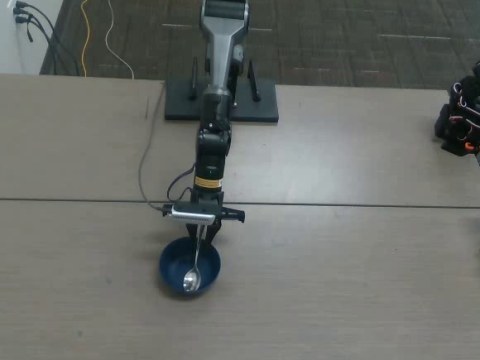} \\
\end{tabular}
\caption{\textbf{Successful real-robot rollouts of \method} (front camera; frames left to right in time). From top: placing the can on the plate; one button board under all four instructions, shown as the shared start frame and the final frame for each instruction; stacking the bowls; putting the spoon in the bowl.}
\label{fig:real-success}
\end{figure}

\begin{figure}[p]
\centering
\newcommand{\rrimg}[1]{\includegraphics[width=0.2\textwidth]{fig/real_rollouts/#1}}
\newcommand{\rrlab}[1]{\rotatebox{90}{\parbox{0.15\textwidth}{\centering\scriptsize\sffamily #1}}}
\setlength{\tabcolsep}{0.6pt}\renewcommand{\arraystretch}{0.5}
\begin{tabular}{@{}>{\centering\arraybackslash}m{0.03\textwidth}*{4}{>{\centering\arraybackslash}m{0.2\textwidth}}@{}}
\rrlab{Missed grasp} & \rrimg{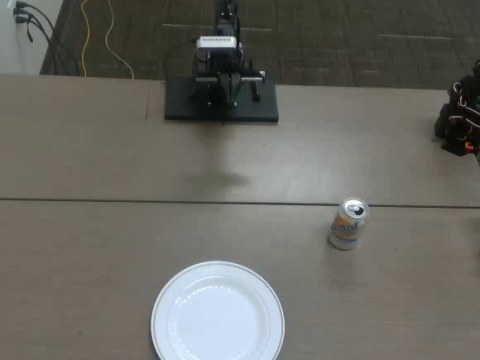} & \rrimg{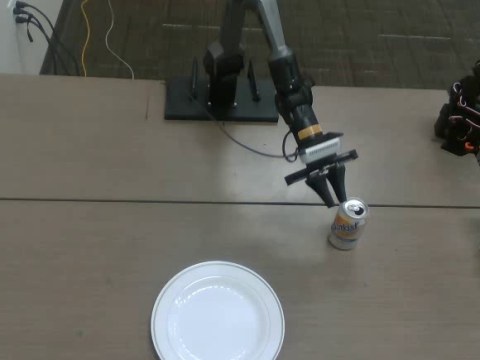} & \rrimg{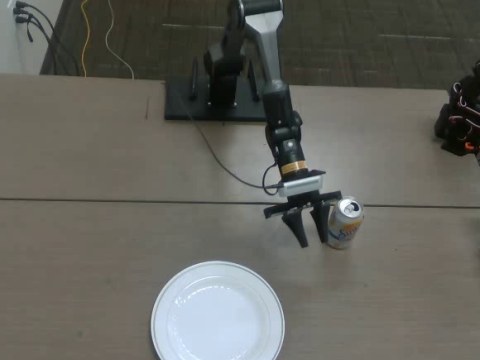} & \rrimg{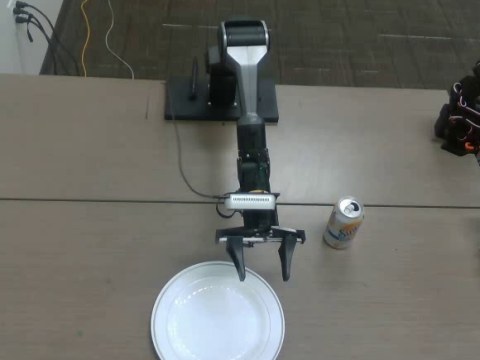} \\
\rrlab{Knocked over} & \rrimg{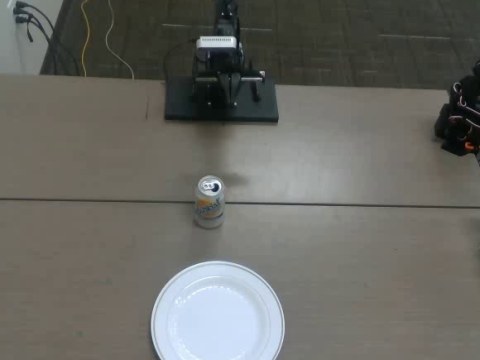} & \rrimg{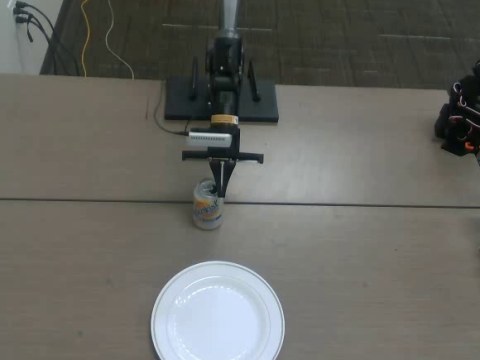} & \rrimg{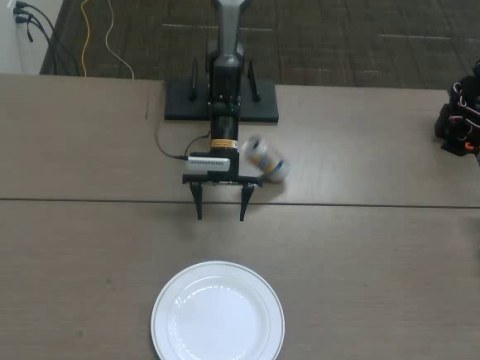} & \rrimg{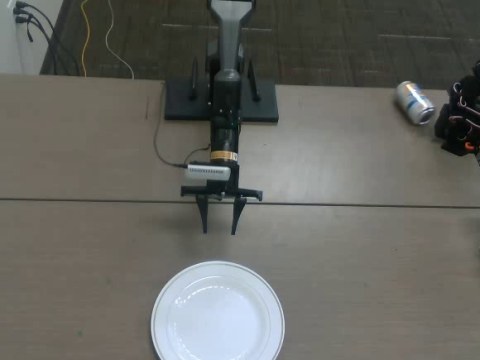} \\
\rrlab{Pushed away} & \rrimg{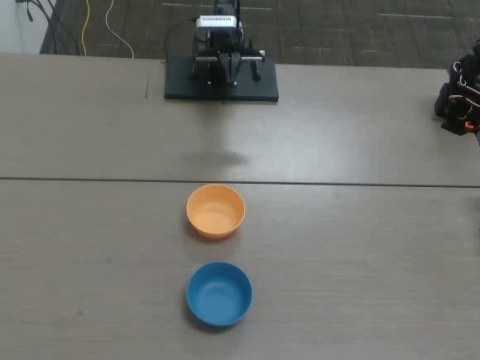} & \rrimg{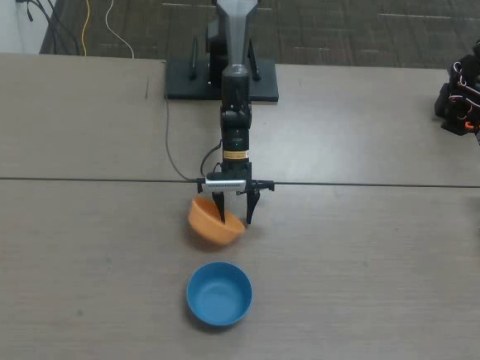} & \rrimg{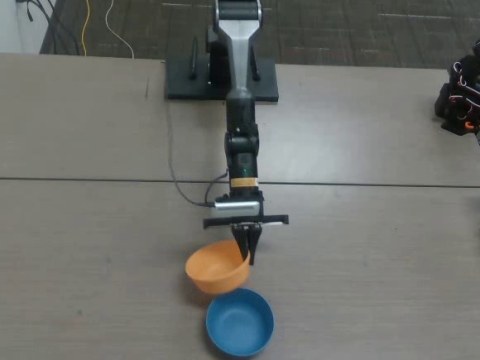} & \rrimg{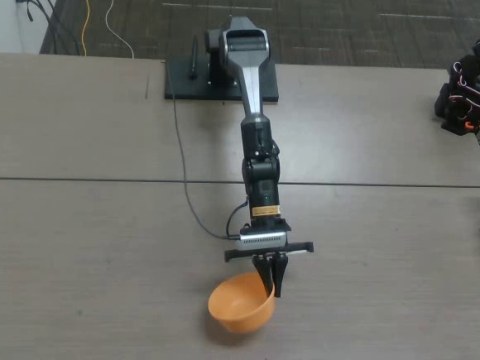} \\
\rrlab{Released off target} & \rrimg{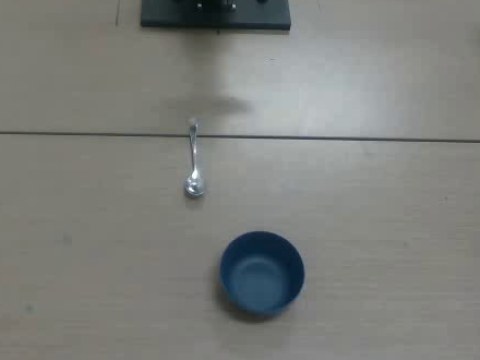} & \rrimg{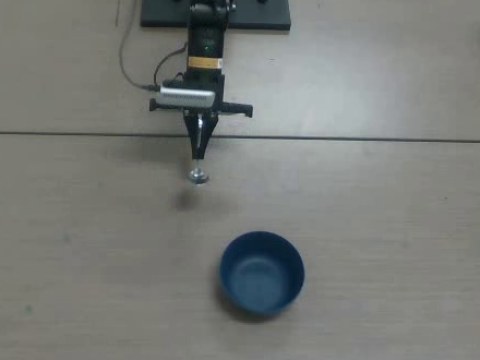} & \rrimg{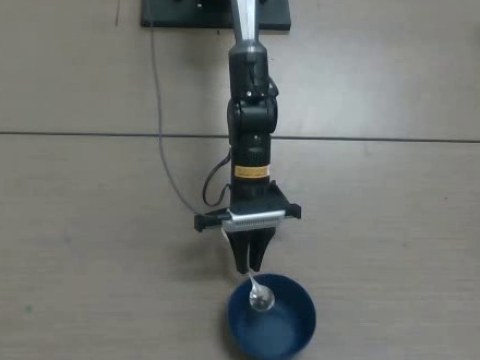} & \rrimg{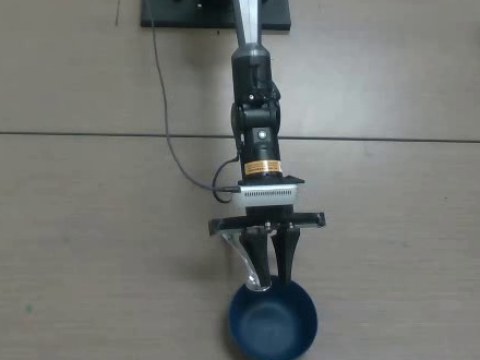} \\
\end{tabular}
\caption{\textbf{Typical failures of \method} (front camera; frames left to right in time). Missed grasp: the fingers close beside the can, and the arm moves on to the plate without it. Knocked over: the can tips over as the fingers close and rolls away. Pushed away: the carried bowl is lowered onto the rim of the other bowl and pushes it out of view. Released off target: the gripper opens above the bowl, but the spoon stays hooked on one finger.}
\label{fig:real-failures}
\end{figure}

\paragraphx{Setup.}
We use an AgileX PiPER 6-DoF arm with a parallel gripper, observed by a front and a wrist camera (Intel RealSense D435i; \cref{fig:real}a), and control it in joint space with 7-D actions: absolute targets for the six joints and the gripper opening.
We collect 400 teleoperated demonstrations (100 per task, 25 per button color) with object poses randomized inside a $30{\times}40$\,cm workspace.
\method is trained on all four tasks jointly from the same Wan2.1 donor and then distilled into \methodE; Fast-WAM$^{*}$ is fine-tuned from its released weights on the same data.
Each method is evaluated on 80 trials (20 per task); before every trial the arm returns to the same home pose and the objects are placed by hand at new positions.
A trial succeeds if the instructed end state is reached within 60\,s.

\paragraphx{Results (\cref{tab:real}).}
\method matches Fast-WAM$^{*}$ at 71.3\% success while reducing memory from 12.7 to 3.7\,GB.
\methodE achieves 68.8\% and 67.5\% success at $(4,2)$ and $(2,2)$, respectively, while reducing latency from 160 to 58 and 46\,ms.

\paragraphx{Failure analysis.}
\Cref{fig:real-success} shows a successful rollout of \method for each task.
Of its 23 failures, 19 occur in the final contact phase (\cref{fig:real-failures}): the gripper closes beside or above the object (10), knocks it over or pushes it away (4), or drops it or releases it off target (5).
After a missed grasp, the policy usually re-tries from almost the same pose until the trial ends, but sometimes moves on to the plate or bowl with an empty gripper, i.e., it does not notice that the grasp failed.
The remaining 4 failures are not contact errors: the arm reaches the instructed button but slides onto the one behind it (2), or hovers above the object without grasping (2).
Failures are most frequent when the object stands close to the robot base.

\section{Conclusion}
\label{sec:conclusion}

\method shows that a world action model does not need 30 distinct layers per expert.
Storing 12 blocks per expert and looping a shared core gives a 0.58B-parameter policy that is competitive with world action models ten times larger on LIBERO and RoboTwin~2.0.
Because inference cost is set by the number of blocks executed per query, the decisive step is to make that number adjustable: decoupled, late-aligned video and action loops and elastic self-distillation let one checkpoint run anywhere between 24.7 and 86.2\,ms with at most one point change in LIBERO success.
Our controlled study points to where the remaining gains lie: video depth matters more than action depth, action depth must be trained to be removable, and a stronger teacher yields a stronger elastic student.

{\small
\bibliographystyle{plainnat}
\bibliography{main}
}

\clearpage
\appendix
\counterwithin{table}{section}
\counterwithin{figure}{section}
\renewcommand{\thetable}{\Alph{section}\arabic{table}}
\renewcommand{\thefigure}{\Alph{section}\arabic{figure}}

\section{Re-evaluated LIBERO baselines}
\label{sec:app-eval}

Every model, including the re-evaluated baselines, is run by one evaluator under the protocol of \cref{sec:setup}.
Each model family runs its own native inference; \methodE uses the $N$ of each setting.
\Cref{tab:app-rerun} compares our re-evaluations with the averages reported by the original papers.

\begin{table}[H]
\centering
\caption{\textbf{Re-evaluated LIBERO baselines}: average success (\%) reported by each paper and obtained under our protocol with the released checkpoints (6{,}000 episodes).}
\label{tab:app-rerun}
\tabfont
\setlength{\tabcolsep}{9pt}
\begin{tabular}{@{}lcc@{}}
\toprule
Method & Reported & Ours \\
\midrule
Light-WAM~\cite{li2026light} & 97.2 & 95.77 \\
Fast-WAM~\cite{yuan2026fast} & 97.6 & 96.63 \\
Faster-WAM~\cite{zhao2026faster} & 99.0 & 98.72 \\
\bottomrule
\end{tabular}
\end{table}

\section{Additional elastic-inference results}
\label{sec:app-elastic}

\Cref{tab:app-elastic-short} lists the elastic student distilled from the shorter-schedule model, which is the data of \cref{fig:elastic}b, together with its teacher and the specialist, which uses the same recipe with every update at $(K_a,N)=(1,2)$.
\Cref{tab:app-naive} gives the data of \cref{fig:elastic}a: the fixed 4/4 model and \methodE at test-time loop counts, including loop counts that were not used in training.

\begin{table}[H]
\centering
\caption{\textbf{Elastic student on the shorter schedule} (LIBERO success \%), the specialist distilled only for $(1,2)$, and the teacher; data of \cref{fig:elastic}b.}
\label{tab:app-elastic-short}
\tabfont
\setlength{\tabcolsep}{7pt}
\begin{tabular}{@{}cccccccc@{}}
\toprule
$K_a$ & $N$ & Lat.\,(ms) & Spatial & Object & Goal & Long & Avg. \\
\midrule
\tabrows{tables/a_elastic_short_rows}
\bottomrule
\end{tabular}
\end{table}

\begin{table}[H]
\centering
\caption{\textbf{Running with $K_a$ action loops at test time} ($N{=}10$, shorter schedule): the fixed 4/4 model (trained with $K_a{=}4$: 94.92\%) and \methodE at loop counts it was not trained with; data of \cref{fig:elastic}a. \methodE at its trained loop counts is in \cref{tab:app-elastic-short}.}
\label{tab:app-naive}
\tabfont
\setlength{\tabcolsep}{8pt}
\begin{tabular}{@{}cccccc@{}}
\toprule
$K_a$ & Spatial & Object & Goal & Long & Avg. \\
\midrule
\multicolumn{6}{@{}l}{\textit{Fixed 4/4 model}} \\
\tabrows{tables/a_naive_rows}
\midrule
\multicolumn{6}{@{}l}{\textit{\methodE}} \\
\tabrows{tables/a_elastic_untrained_rows}
\bottomrule
\end{tabular}
\end{table}

\section{Ablations per suite}
\label{sec:app-ablations}

\Cref{tab:app-loops,tab:app-init,tab:app-part} give the per-suite results behind \cref{tab:ablations}.

\paragraphx{Where shallow settings fail.}
Reducing depth costs little on LIBERO-Spatial but much on LIBERO-Long (\cref{tab:app-loops,tab:app-init,tab:app-part}): Long falls from 87.87\% (4/4) to 87.07\% (4/2), 86.27\% (4/1), 82.07\% (2/2), and 78.20\% (1/4), below Dense-12 (79.73\%) for 1/4, and every initialization and partition alternative also loses most on Long, except 6/6/0, which loses 6.1 points on Object and 5.8 on Long.

\begin{table}[H]
\centering
\caption{\textbf{Loop allocation and dense controls} on the shorter schedule (LIBERO per suite and RoboTwin-10). Latency: one H100, batch 1. KV-mix lets the single action loop attend to a mixture of all video loops.}
\label{tab:app-loops}
\tabfont
\setlength{\tabcolsep}{5pt}
\begin{tabular}{@{}lcccccccccc@{}}
\toprule
 & & & & & \multicolumn{5}{c}{LIBERO} & \\
\cmidrule(lr){6-10}
Model & $K_v$ & $K_a$ & Params & Lat.\,(ms) & Spatial & Object & Goal & Long & Avg. & RT-10 \\
\midrule
\tabrows{tables/a_loops_rows}
\bottomrule
\end{tabular}
\end{table}

\begin{table}[H]
\centering
\caption{\textbf{Initialization of the video core} (LIBERO success \%, shorter schedule). Prelude and coda copy Wan2.1 blocks 1--3 and 28--30.}
\label{tab:app-init}
\tabfont
\setlength{\tabcolsep}{8pt}
\begin{tabular}{@{}lccccc@{}}
\toprule
Core & Spatial & Object & Goal & Long & Avg. \\
\midrule
\tabrows{tables/a_init_rows}
\bottomrule
\end{tabular}
\end{table}

\begin{table}[H]
\centering
\caption{\textbf{Partition} $P/C/E$ with $K$ loops (LIBERO success \%, shorter schedule); donor blocks given as prelude / core / coda.}
\label{tab:app-part}
\tabfont
\setlength{\tabcolsep}{7pt}
\begin{tabular}{@{}lclccccc@{}}
\toprule
$P/C/E$ & $K$ & Donor blocks & Spatial & Object & Goal & Long & Avg. \\
\midrule
\tabrows{tables/a_part_rows}
\bottomrule
\end{tabular}
\end{table}

\section{RoboTwin-10}
\label{sec:app-rt10}

RoboTwin-10 contains ten tasks of graded difficulty (\cref{tab:app-rt10}).
Three are solved by nearly every configuration, and the differences concentrate in four contact-rich tasks, whose average is 54.50\% for Dense-12, 55.88\% and 61.00\% for 2/2 and 1/4, and 62.63--66.25\% for the three models with four video loops; Dense-30 reaches 65.25\%. Single tasks vary more than the averages: on \texttt{open\_microwave} the models range from 45.0\% (4/4) to 77.5\% (4/2).
With 2{,}000 episodes per model, each overall average has a 95\% interval of $\pm$1.7--1.8 points.

\begin{table}[H]
\centering
\caption{\textbf{RoboTwin-10 per task} (success \%, average of clean and randomized, 200 episodes per task) for each configuration ($K_v/K_a$).}
\label{tab:app-rt10}
\tabfont
\setlength{\tabcolsep}{6pt}
\begin{tabular}{@{}lccccccc@{}}
\toprule
Task & Dense-12 & 1/4 & 2/2 & 4/1 & 4/2 & 4/4 & Dense-30 \\
\midrule
\tabrows{tables/a_rt10_tasks_rows}
\bottomrule
\end{tabular}
\end{table}

\section{Per-task results}
\label{sec:app-results}

\Cref{tab:app-rt50} lists all 50 RoboTwin~2.0 tasks, and \cref{tab:app-gr1-tasks} all 24 RoboCasa-GR1 tasks for \method and its elastic student.
\Cref{tab:gr1-groups} breaks the RoboCasa-GR1 result down by task group: \method is strongest on the cutting-board and plate tasks and weakest on the placemat tasks.

\begin{table}[H]
\centering
\caption{\textbf{RoboCasa-GR1 by task group} (success \%; number of tasks in parentheses in \cref{tab:app-gr1-tasks}). Roll.: rollouts per task (--: not stated). Prior methods: means of the per-task results in their papers (GR00T-N1.6: as listed by StarVLA); \methodE: $(K_a,N)$.}
\label{tab:gr1-groups}
\tabfont
\setlength{\tabcolsep}{6pt}
\begin{tabular}{@{}lccccccc@{}}
\toprule
Method & Roll. & \makecell{PnP +\\close} & \makecell{Cutting\\board} & \makecell{Place-\\mat} & Tray & Plate & Avg. \\
\midrule
\tabrows{tables/a_gr1_groups_rows}
\bottomrule
\end{tabular}
\end{table}

\begin{table}[H]
\centering
\caption{\textbf{RoboTwin~2.0 per task} for \method (successes out of 100 in the clean and randomized settings, and their average in \%).}
\label{tab:app-rt50}
\tabfontsmall
\setlength{\tabcolsep}{5pt}
\begin{tabular}{@{}lccc@{\hspace{20pt}}lccc@{}}
\toprule
Task & Clean & Rand. & Avg. & Task & Clean & Rand. & Avg. \\
\midrule
\tabrows{tables/a_rt50_tasks_rows}
\bottomrule
\end{tabular}
\end{table}

\begin{table}[H]
\centering
\caption{\textbf{RoboCasa-GR1 per task} for \method and \methodE at $(4,2)$ and $(2,2)$ (success \%, 50 rollouts per task).}
\label{tab:app-gr1-tasks}
\tabfontsmall
\setlength{\tabcolsep}{8pt}
\begin{tabular}{@{}lccc@{}}
\toprule
Task & \method & \methodE $(4,2)$ & \methodE $(2,2)$ \\
\midrule
\tabrows{tables/a_gr1_tasks_rows}
\bottomrule
\end{tabular}
\end{table}

\end{document}